\PassOptionsToPackage{table}{xcolor}
\documentclass{article} %
\usepackage[final]{colm2026_conference}

\usepackage{microtype}
\usepackage{hyperref}
\usepackage{url}
\usepackage{booktabs}
\usepackage{amsmath,amssymb}
\usepackage{iftex}
\ifPDFTeX
\usepackage{CJKutf8}
\fi
\usepackage{xcolor}
\usepackage{graphicx}
\usepackage{enumitem}
\usepackage{wrapfig}
\usepackage{makecell}
\usepackage{tcolorbox}
\usepackage{caption}
\usepackage{adjustbox}
\usepackage{lineno}
\definecolor{darkblue}{rgb}{0, 0, 0.5}
\definecolor{revisionpurple}{HTML}{000000}
\newcommand{\rev}[1]{\textcolor{revisionpurple}{#1}}
\hypersetup{colorlinks=true, citecolor=darkblue, linkcolor=darkblue, urlcolor=darkblue}

\title{Ideological Bias in LLMs' Economic Causal Reasoning}

\author{
Donggyu Lee$^{1}$ \quad
Hyeok Yun$^{2}$ \quad
Jungwon Kim$^{2}$ \quad
Junsik Min$^{3}$ \\
\textbf{Sungwon Park}$^{3\,*}$ \quad
\textbf{Sangyoon Park}$^{4\,*}$ \quad
\textbf{Jihee Kim}$^{2\,*}$ \\[0.6em]
$^{1}$Graduate School of Data Science, KAIST, Daejeon, South Korea \\
$^{2}$College of Business, KAIST, Daejeon, South Korea \\
$^{3}$School of Computing, KAIST, Daejeon, South Korea \\
$^{4}$Division of Social Science, HKUST, Hong Kong, China \\[0.3em]
\texttt{donggyu.lee@kaist.ac.kr}, \texttt{ed\_yun98@kaist.ac.kr}, \\
\texttt{jungwonkim126@kaist.ac.kr}, \texttt{junshik1211@kaist.ac.kr}, \\
\texttt{psw0416@kaist.ac.kr}, \texttt{sangyoon@ust.hk}, \texttt{jiheekim@kaist.ac.kr}
\thanks{$^{*}$Co-corresponding authors.}
}

\newcommand{\mypara}[1]{\paragraph{{\fontencoding{OT1}\fontfamily{ppl}\fontseries{b}\selectfont #1.}} }
\newcommand{\cutsectionup}{\vspace*{-0.12in}}
\newcommand{\cutsubsectionup}{\vspace*{-0.12in}}
\newcommand{\cutparagraphup}{\vspace*
{-0.15in}}
\newcommand{\cutcaptionup}{\vspace*{-0.12in}}

\begin{document}
\ifPDFTeX
\begin{CJK*}{UTF8}{mj}
\fi

\ifcolmsubmission
\linenumbers
\fi

\maketitle

\begin{abstract}
Do large language models (LLMs) exhibit systematic ideological bias when reasoning about economic causal effects? As LLMs are increasingly used in policy analysis and economic reporting, where directionally correct causal judgments are essential, this question has direct practical stakes. We present a systematic evaluation by extending the EconCausal benchmark with ideology-contested cases—instances where intervention-oriented (pro-government) and market-oriented (pro-market) perspectives predict divergent causal signs. From 10,490 causal triplets (treatment–outcome pairs with empirically verified effect directions) derived from top-tier economics and finance journals, we identify 1,056 ideology-contested instances and evaluate 20 state-of-the-art LLMs on their ability to predict empirically supported causal directions. We find that ideology-contested items are consistently harder than non-contested ones, and that across \rev{17} of 20 models, accuracy is systematically higher when the empirically verified causal sign aligns with intervention-oriented expectations than with market-oriented ones. Moreover, when models err, their incorrect predictions disproportionately lean intervention-oriented—and this directional skew is not eliminated by one-shot in-context prompting. These results highlight that LLMs are not only less accurate on ideologically contested economic questions, but systematically less reliable in one ideological direction than the other, underscoring the need for direction-aware evaluation in high-stakes economic and policy settings.
\end{abstract}

\cutsectionup
\section{Introduction}
\label{sec:introduction}
\cutsectionup
 Do Large Language Models exhibit systematic ideological bias when reasoning about economic causal effects? This question is pressing because LLMs are increasingly deployed in high-stakes analytical workflows, such as economic reporting, policy evaluation, and corporate decision support \citep{kwon2024large, ludwig2025large, handler2024large}, where predicting causal directions correctly is essential. Yet in many such settings, that direction is genuinely contested: a single intervention can trigger competing mechanisms whose relative magnitudes are debated along ideological lines. A minimum wage increase, for instance, is expected to reduce hiring under a market-oriented framework that prioritizes labor cost effects, but to sustain employment under an intervention-oriented framework that emphasizes demand or job-matching gains. When LLMs predict causal directions in such settings, their outputs may implicitly privilege one set of priors over another, producing an invisible tilt in downstream policy advice.

 Existing approaches to detecting such effects fall short in complementary ways. Work on LLM bias typically relies on survey-style instruments that probe expressed opinions or preferences, often in social or political domains, and does not assess how models reason about causal relationships grounded in empirical evidence. In contrast, benchmarks for causal reasoning focus on formal logic or context-dependent inference, but assume a single correct expectation for each causal effect and do not account for settings where competing, internally consistent frameworks imply different expected directions. As a result, we lack a framework for asking whether LLMs reason about economic causation more reliably in one ideological direction than another, the setting where bias would have the most direct policy consequences.
 
 In this paper, we present the first systematic analysis of ideological bias in LLMs' economic causal reasoning. Leveraging the EconCausal benchmark dataset—a large-scale corpus of 10,490 context-annotated causal triplets derived from top-tier economics and finance journals—we identify 1,056 ideologically contested triplets (10.1\%) where \rev{market-oriented frameworks (emphasizing market allocation and limited government intervention) and intervention-oriented frameworks (emphasizing government action to address market failures)} predict divergent causal signs. We evaluate LLM accuracy separately across contested and non-contested settings, and define a directional bias measure that captures systematic asymmetries in model errors.

Through an extensive evaluation of 20 state-of-the-art LLMs, we document a striking asymmetry in reasoning ability. Models exhibit lower accuracy on ideologically contested triplets and perform significantly better when the empirical truth aligns with intervention-oriented expectations, showing an average accuracy gap of \rev{+8.5} to \rev{+14.2} percentage points relative to market-oriented cases. Errors are not random but systematically directional, and this pattern is robust across model families and difficulty levels. These findings highlight substantial reliability risks when LLMs serve as intermediaries in policy analysis, where implicit ideological anchoring can lead to systematically skewed conclusions.

Our contributions are as follows. First, we introduce the notion of ideologically contested causal relationships as a new lens for studying LLM causal reasoning beyond context dependence. Second, we construct a labeled subset of EconCausal distinguishing contested from non-contested triplets, enabling controlled evaluation across these settings. Third, we provide a systematic empirical analysis showing lower accuracy and asymmetric performance in contested cases. Fourth, we define a directional bias measure and document consistent asymmetry patterns across a range of models. Together, our results highlight the importance of evaluating causal reasoning not only across contexts, but also across settings where disagreement over causal expectations is intrinsic.

\begin{figure}[tb]
    \centering
    \includegraphics[width=1\linewidth]{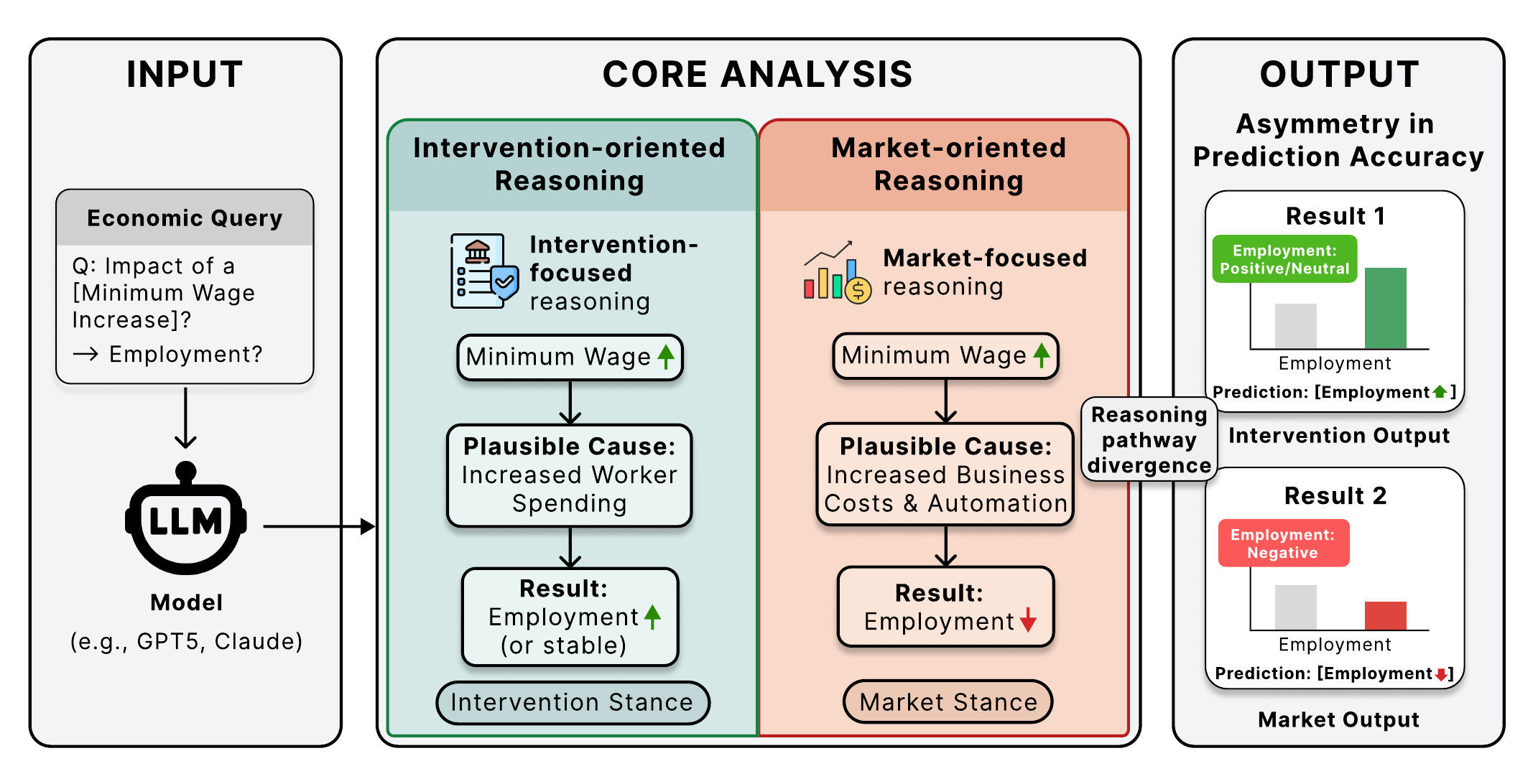}
    \cutcaptionup
    \caption{Overview of ideological divergence in LLM economic causal reasoning.}
    \label{fig:overview}
\end{figure}%

\cutsectionup
\section{Related Work}
\vspace{-3mm}
\subsection{Political and Ideological Bias in LLMs}
\cutsubsectionup

Whether LLMs exhibit systematic political biases has attracted considerable attention. A common approach applies political orientation instruments to LLMs and positions their responses on ideological scales, and the emerging consensus is that instruction-tuned LLMs lean left on social and cultural dimensions \citep{ santurkar2023, rozado2024, hartmann2023, motoki2023}. More recent efforts confirm a leftward tendency but show that its magnitude was overstated by earlier instruments, and that small changes in question format can substantially shift measured positions \citep{rottger2024, faulborn2025}.

These leanings have been traced to multiple sources. \citet{feng2023} show that political biases in pretraining corpora propagate into downstream task predictions. \citet{fulay2024} find that optimizing for truthfulness amplifies left-leaning tendencies in larger models, and \citet{exler2025} document a monotonic relationship between parameter count and alignment with left-leaning parties. %

However, these studies focus on expressed preferences in response to opinion-like questions. They reveal what positions LLMs appear to endorse, but not whether ideological priors systematically distort reasoning about verifiable factual claims. \rev{This distinction is consequential in economic settings because users may interpret a causal-sign prediction as a summary of empirical evidence even when it partly reflects a prior absorbed during training.} Whether ideological leanings also shape how LLMs reason about economic causal relationships---where ground-truth effect signs are available from peer-reviewed empirical research---remains unexplored.

\cutsubsectionup
\subsection{LLMs in Economic and Causal Reasoning}
\cutsubsectionup
Work on causal reasoning suggests that LLMs often rely on priors rather than principled inference, which may help explain why directional bias can arise in economic prediction. Benchmarks such as CLadder and CORR2CAUSE consistently find that LLMs struggle with genuine causal reasoning across different levels of formalism \citep{jin2023cladder,jin2024corr2cause,wang2024causalbench,chen2024calm, yu2025causaleval}. A common explanation is that their causal judgments rely heavily on priors absorbed during pretraining rather than principled inference: LLMs have been described as ``causal parrots,'' shown to exploit textual metadata rather than observational evidence, to adopt rigid rule-like strategies, and to inadequately revise beliefs when confronted with contradictory evidence \citep{zecevic2023,kiciman2023,wilie2024, dettki2026}.

Related work on LLMs in economic and policy settings points to similar limitations. LLMs have already been applied to policy-adjacent tasks such as classifying central bank communications \citep{silva2025}, while existing benchmarks show persistent weaknesses in economic and causal reasoning \citep{raman2024steer,raman2025steerme,quan2024,guo2024}. Yet this literature evaluates performance mainly in terms of accuracy and robustness, not whether prediction errors are systematically directional along ideological lines. Our work addresses this gap by testing whether LLMs exhibit ideologically asymmetric errors when predicting economic causal effect signs documented in the empirical literature.

\cutsectionup
\section{EconCausal and Ideology-contested Extension}
\label{sec:data}
\cutsectionup
\subsection{EconCausal Benchmark Overview}
\cutsubsectionup
\label{subsec:econcausal_overview}
This study leverages the experimental framework of the EconCausal benchmark \citep{lee2026econcausalcontextawarecausalreasoning}, which provides a dataset of causal relationships extracted from empirical studies in top-tier economics and finance journals.
For each data instance, a causal triplet $(T, O, Sign)$ consisting of the treatment $T$, outcome $O$, and the directional sign of the effect $Sign$ was extracted, along with a corresponding context paragraph $C$ that provides the essential socio-economic background for the observed relationship.

The primary experimental setup adopts a sign prediction task, requiring language models to predict the direction of the causal effect $Sign$ given a context $C$ and a treatment-outcome pair $(T, O)$ as inputs. The sign label space is defined as
\[
\mathcal{S} = \{+, -, \text{None}, \text{Mixed}\},
\]
where the four classes represent significantly positive, significantly negative, insignificant, and heterogeneous/complex effects, respectively.

\subsection{Defining Ideology-contested Causal Triplets}
\label{subsec:ideology_definition}
\cutsubsectionup

Following the EconCausal benchmark, each of the {10{,}490} causal triplets is assigned a ground-truth sign \rev{$Sign$} based on the original authors' empirical specification.
To construct ideology-conditioned labels, we queried four LLMs (GPT, Claude, Qwen, and Grok) once per triplet, asking each model to jointly assign an intervention-oriented expected sign and a market-oriented expected sign while remaining blinded to the paper’s empirical result. Detailed prompt is provided in Appendix ~\ref{app:prompts}.

These signs are defined along the economic dimension of ideology \citep{alesina2011preferences, feldman2014understanding}: $s^{\mathrm{intervention}}$ reflects the expected causal sign under an intervention-oriented perspective that favors active government intervention to correct market failures, reduce inequality, and expand social insurance, while $s^{\mathrm{market}}$ reflects expectations under a market-oriented perspective that favors market-determined outcomes, individual responsibility, and limited government intervention. \rev{Both ideology-conditioned signs belong to $\mathcal{S}$ and are inferred independently of $Sign$; they identify disagreement between priors but do not determine the benchmark answer.}
\rev{We retained a triplet for ideology-contested labeling only when at least
three of the four models provided valid sign pairs. We then took strict
majority votes separately for the intervention- and market-oriented signs. A
triplet was classified as ideology-contested when these two final votes
differed.}

Formally, the ideology-contested subset is defined as
{\color{revisionpurple}
\begin{equation}
\mathcal{D}_{\mathrm{cont}}
=
\left\{
  x \in \mathcal{D}
  \;\middle|\;
  s^{\mathrm{intervention}}(x) \neq s^{\mathrm{market}}(x)
\right\}.
\end{equation}
}

This procedure yields {1{,}056} ideology-contested triplets (10.1\% of the full set).
\rev{For directional accuracy comparisons, we use the subset
$\mathcal{D}_{\mathrm{dir}}=\{x\in\mathcal{D}_{\mathrm{cont}}:Sign(x)\in\{s^{\mathrm{intervention}}(x),s^{\mathrm{market}}(x)\}\}$. Among the 1,056 contested triplets, 878 belong to $\mathcal{D}_{\mathrm{dir}}$:}
\begin{itemize}
  \item \textbf{Intervention-truth} (\rev{507} triplets, \rev{57.7\%}): the ground-truth sign $Sign$ matches $s^{\mathrm{intervention}}$.
  \item \textbf{Market-truth} (\rev{371} triplets, \rev{42.3\%}): the ground-truth sign $Sign$ matches $s^{\mathrm{market}}$.
\end{itemize}

\cutparagraphup
\paragraph{Expert Validation.}
\rev{Two economics professors independently reviewed a representative
subset of 126 triplets (18 per subcategory $\times$ 7 subcategories,
balanced across intervention-oriented, market-oriented, and neutral alignments). For each item, they reviewed the original causal triplet together with the ensemble's ideology classification and made a yes/no judgment about whether that classification was valid; they were not re-labeling $Sign$. The experts accepted 93.60\% and 98.35\% of the classifications, respectively, and the reported 93.33\% inter-rater figure is their raw agreement on this yes/no validity judgment.}
The subfield composition of the ideology-contested subset, based on
JEL classification codes, is detailed in Appendix~\ref{app:subfield}.

\cutsectionup
\section{Main Results: Ideological Distortion in Economic Causal Reasoning}
\label{sec:main_results}

\cutsectionup
\subsection{Evaluation Settings}
\label{subsec:models}
\cutsubsectionup
We evaluate a diverse set of LLMs spanning both closed-source and open-source families.
The closed-source group includes OpenAI GPT, Anthropic Claude, Google Gemini, and xAI Grok;
the open-source group includes Meta Llama and Alibaba Qwen, chosen across varying parameter scales to examine the effect of model size.

\rev{As a prompt-robustness check, we evaluate five prompt variants over three
runs across 16 models on the corrected 878-item directional subset. The
intervention--market accuracy gap remains positive under every variant, ranging
from 7.4 to 9.6~pp for closed-source models and from 13.1 to 14.6~pp for
open-source models (Table~\ref{tab:prompt_ablation_corrected878} in
Appendix~\ref{app:prompt_robustness}).}

For evaluation, we measure overall accuracy as well as accuracy on the ideology-contested and non-contested subsets separately.
Within the ideology-contested subset, we define intervention-truth accuracy ($\text{Acc}^{\text{intervention}}$) as the accuracy on items whose ground truth aligns with intervention expectations, and market-truth accuracy ($\text{Acc}^{\text{market}}$) as the accuracy on items whose ground truth aligns with market expectations.
We further define $\text{Errors}^{\text{intervention}}$ and $\text{Errors}^{\text{market}}$ as the numbers of incorrect predictions whose predicted sign aligns with intervention and market expectations, respectively, and let $\text{Errors}^{\text{total}}$ denote the total number of errors \rev{among the 878 directionally aligned items}.
Their accuracy gap and directional error bias are then formalized as:
\begin{equation}
  \Delta_{\text{acc}} = \text{Acc}^{\text{intervention}} - \text{Acc}^{\text{market}}, \qquad
  B_{\text{dir}} = \rev{100\times}\frac{\text{Errors}^{\text{intervention}} - \text{Errors}^{\text{market}}}{\text{Errors}^{\text{total}}}
\end{equation}
\rev{The directional accuracies are computed on 507 intervention-truth and 371 market-truth items, respectively. Positive $\Delta_{\text{acc}}$ indicates higher accuracy on intervention-truth items, while positive $B_{\text{dir}}$, reported in percentage units, indicates intervention-leaning errors.}

\subsection{Main Results: Performance and Bias on Ideology-contested Items}
\cutsubsectionup
\label{subsec:main-results}

\begin{figure}[tb]
    \centering
    \includegraphics[width=0.9\linewidth]{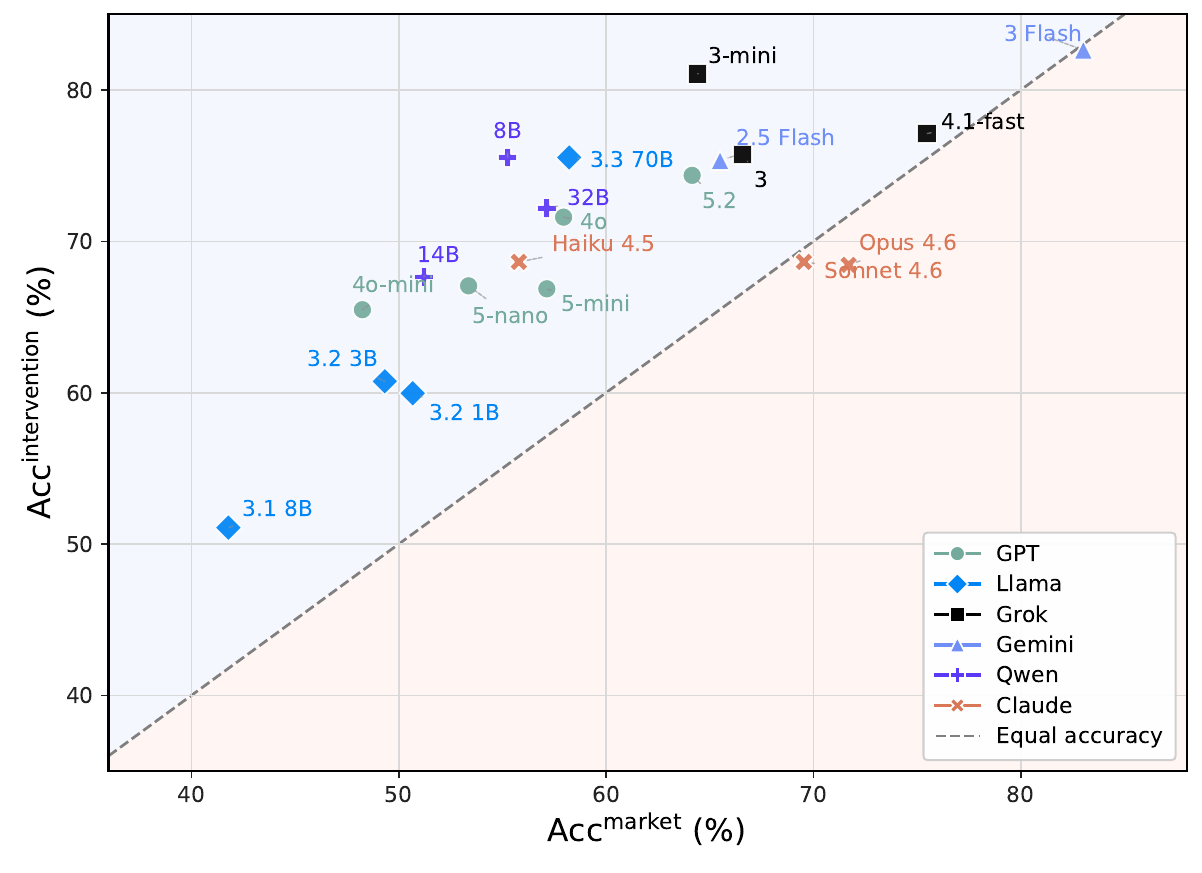}
    \cutcaptionup
    \caption{Accuracy on intervention-truth versus market-truth items across 20 LLMs. \rev{The diagonal denotes equal accuracy ($\Delta_{\mathrm{acc}}=0$); points above it have higher intervention-truth accuracy.}}
    \label{fig:main_results}
\end{figure}%

Before examining accuracy, we note that the ground-truth labels themselves lean toward intervention-side (\rev{57.7\%} vs.\ \rev{42.3\%} among directional items), and the models' average predictions closely track this base rate (\rev{59.5\%} vs.\ \rev{40.5\%}) \rev{when restricted to predictions matching one of the two ideology-conditioned signs}. Figure~\ref{fig:main_results} reports model-level accuracy on the \rev{878-item directionally aligned subset}, focusing on intervention-truth versus market-truth items. The full table of main results is reported in Appendix~\ref{app:full_model}. We organize the findings along three dimensions: overall difficulty, asymmetric accuracy, and directional error bias.

\cutparagraphup
\mypara{Ideology-contested triplets are harder} Across all 20 models, accuracy on the ideology-contested subset is consistently lower than on the non-contested remainder.
Closed-source models average \rev{60.8\%} on \rev{all 1,056 ideology-contested} items versus 74.5\% on non-contested items, a gap of \rev{13.7} percentage points~(pp).
Open-source models show a comparable pattern at a lower absolute level, averaging \rev{51.7\%} versus 63.8\% (\rev{12.1}~pp gap).
This uniform degradation indicates that economically contested causal relationships pose a systematically harder challenge for LLMs, independent of model family or scale.
\cutparagraphup
\mypara{Accuracy is asymmetric across ideological directions}
Closed-source models average \rev{72.5\%} on intervention-truth items but only \rev{64.1\%} on market-truth items ($\Delta_{\text{acc}} = \rev{+8.5}$~pp); open-source models show a wider gap at $\rev{+14.2}$~pp.
Of the 20 models, \rev{17} exhibit a positive $\Delta_{\text{acc}}$, with the largest gaps exceeding \rev{17}~pp (GPT-4o-mini, Qwen~3-8B, Llama~3.3-70B).
The Claude family is a notable exception: Sonnet~4.6 and Opus~4.6 display a slight market-truth advantage ($\Delta_{\text{acc}} = \rev{-0.9}$ and $\rev{-3.3}$, respectively).
\cutparagraphup
\mypara{Errors are systematically directional} The accuracy asymmetry is corroborated by the directional bias score $B_{\text{dir}}$.
Closed-source models show a modest intervention-leaning bias on average ($B_{\text{dir}} = \rev{+2.3}$), while open-source models exhibit a substantially stronger intervention skew ($B_{\text{dir}} = \rev{+7.3}$).
Among the 20 models, 15 exhibit $B_{\text{dir}} > 0$, with \rev{large} directional biases observed in Llama~3.3-70B ($B_{\text{dir}} = \rev{+12.5}$), Qwen~3-8B ($B_{\text{dir}} = \rev{+16.9}$), and GPT-4o-mini ($B_{\text{dir}} = \rev{+6.3}$).
The consistent positive values of $B_{\text{dir}}$ across the majority of models suggest that prediction errors on ideology-contested items are not uniformly distributed but instead reflect a systematic intervention-leaning tendency.
\cutsubsectionup
\subsection{Heterogeneity Across Economic Subfields}
\cutsubsectionup
To examine the fine-grained effects of ideological bias, we divide the directionally aligned subset into seven major economic subfields based on JEL (Journal of Economic Literature) classification codes, a standardized taxonomy maintained by the American Economic Association that assigns hierarchical alphanumeric labels to economic research topics~\citep{jel2024}. Detailed mapping rules and subfield-level accuracy gaps are provided in Appendix~\ref{app:subfield}.

An intervention-truth advantage appears in \rev{all} seven subfields but is highly heterogeneous. The largest asymmetries arise in healthcare and welfare/redistribution, while labor and financial regulation exhibit more moderate gaps. \rev{Trade has the smallest positive gap.}
\rev{These results indicate that the strength of the intervention-truth advantage varies substantially by subfield rather than operating uniformly across domains.} The especially large intervention-truth advantage in healthcare ($\Delta_{\text{acc}} = \rev{+20.2}$~pp) and welfare/redistribution ($\Delta_{\text{acc}} = \rev{+14.7}$~pp) is consistent with LLMs responding more asymmetrically in policy domains where state intervention is ideologically salient. At the same time, \rev{the small positive gap in trade} ($\Delta_{\text{acc}} = \rev{+1.0}$~pp) cautions against reducing the pattern to a blanket intervention bias. This heterogeneity clarifies that ideological distortion is concentrated in specific policy-relevant areas of economics rather than evenly distributed across all domains. \looseness=-1

\cutsubsectionup
\subsection{Robustness Check: Difficulty Matching}
\label{sec:robustness}
\cutsubsectionup

\begin{table}[h!]
\centering
\begin{tabular}{lcc}
\toprule
 & Unmatched & Matched \\
 & ($n_{\text{Intervention}}=\rev{507},\;n_{\text{Market}}=\rev{371}$) & (\rev{$n=352$ per side}) \\
\midrule
Acc$_{\text{Intervention}}$ (\%) & \rev{70.3} & \rev{69.0} \\
Acc$_{\text{Market}}$ (\%) & \rev{59.8} & \rev{59.5} \\
$\Delta_{\text{acc}}$ (pp) & \rev{$+$10.5} & \rev{$+$9.5} \\
\rev{Models with $\Delta_{\mathrm{acc}}>0$} & \rev{17\,/\,20} & \rev{17\,/\,20} \\
Models with $B_{\mathrm{dir}}>0$ & \rev{15\,/\,20} & \rev{18\,/\,20} \\
\bottomrule
\end{tabular}
\cutcaptionup
\caption{Accuracy gap between intervention-truth and market-truth items before and after difficulty matching. All accuracy values are 20-model means.}\label{tab:robustness}
\end{table}

A natural concern is that market-truth items may be intrinsically harder, and that the accuracy gap in Figure~\ref{fig:main_results} reflects this difficulty differential rather than ideological distortion. We scored \rev{all 878 directionally aligned triplets} on a 1--5 overall difficulty scale using GPT-5-mini and matched intervention-truth and market-truth triplets one-to-one within each \emph{theme $\times$ difficulty} cell, yielding \rev{352} triplets per side (details in Appendix~\ref{app:difficulty}).

\rev{As Table~\ref{tab:robustness} shows, the accuracy gap persists after
difficulty matching, with the same 17 of 20 models retaining
$\Delta_{\mathrm{acc}}>0$. The observed asymmetry is therefore not fully
explained by the measured difference in item difficulty. In a logistic
regression on all 878 directional items, the intervention-truth coefficient is
0.455 after model, theme, and difficulty fixed effects; adding empirical-sign
fixed effects attenuates it by 10.7\% to 0.406, while it remains positive and
statistically significant (Appendix~\ref{app:difficulty}).}

\cutsectionup
\section{Ideological Steering via In-Context Examples}
\label{sec:icl}
\cutsectionup
 
The preceding analysis established that LLMs \rev{are systematically more accurate on intervention-truth causal triplets}, with an average accuracy gap of $\Delta_{\text{acc}} = \rev{+10.5}$~pp and intervention-leaning error directionality.
In this section, we examine whether in-context examples---one of the simplest prompt-level interventions---can steer model predictions toward a target ideological direction. \looseness=-1
 \cutsubsectionup
\subsection{\rev{Experimental Setup}}
\label{sec:icl_setup}
\cutsubsectionup
\rev{We compare four conditions: no example (\textsc{None}), a non-contested
example (\textsc{Non-Contested}), an intervention-aligned example
(\textsc{Intervention-Ex}), and a market-aligned example
(\textsc{Market-Ex}). The \textsc{None} columns use the corrected-878 baseline
(507 intervention-truth and 371 market-truth items), while the example columns
preserve the original matched experiment. Their target--example-pair counts are
1,325/761 for \textsc{Non-Contested}, 143/135 for
\textsc{Intervention-Ex}, and 158/125 for \textsc{Market-Ex}
(intervention-truth/market-truth). Examples share at least two JEL codes with
the target and have Jaccard similarity of at least $0.5$ over their JEL-code
sets (Appendix~\ref{app:icl_matching}). Our primary metric is the
\emph{Intervention--Market Example Gap},
$\Delta_{\text{example}} = \mathrm{Acc}_{\textsc{Intervention-Ex}}-
\mathrm{Acc}_{\textsc{Market-Ex}}$.}
 \cutsubsectionup
\subsection{Results}
\label{sec:icl_results}
\cutsubsectionup
\begin{table*}[htbp]
\centering

\normalsize
\setlength{\tabcolsep}{4pt}
\renewcommand{\arraystretch}{1.1}

\begin{adjustbox}{max width=\textwidth}
\begin{tabular}{@{}ll ccccc ccccc@{}}
\toprule
& & \multicolumn{5}{c}{\textit{Intervention-Truth Target}} & \multicolumn{5}{c}{\textit{Market-Truth Target}} \\
\cmidrule(lr){3-7} \cmidrule(lr){8-12}
\textbf{Family} & \textbf{Model}
  & None & Non-contested & Intervention-Ex & Market-Ex &  $\Delta_{\text{example}}$
  & None & Non-contested & Intervention-Ex & Market-Ex &  $\Delta_{\text{example}}$ \\
\midrule
\multicolumn{12}{@{}l}{\textit{Closed-Source Models}} \\[2pt]
OpenAI
  & GPT-4o-mini & \rev{65.5} & 47.5 & 50.3 & 58.2 & \color{red}{$-$7.9} & \rev{48.2} & 52.6 & 60.7 & 50.4 & \color{blue}{+10.3} \\
  & GPT-4o & \rev{71.6} & 43.4 & 45.5 & 43.7 & \color{blue}{+1.8} & \rev{58.0} & 46.6 & 52.6 & 63.2 & \color{red}{$-$10.6} \\
  & GPT-5-nano & \rev{67.1} & 55.7 & 55.9 & 53.8 & \color{blue}{+2.1} & \rev{53.4} & 61.6 & 55.6 & 56.0 & \textbf{\color{red}{$-$0.4}} \\
  & GPT-5-mini & \rev{66.9} & 68.7 & 74.1 & 63.9 & \color{blue}{+10.2} & \rev{57.1} & 55.2 & 56.3 & 64.8 & \color{red}{$-$8.5} \\
  & GPT-5.2 & \rev{74.4} & 69.1 & 72.7 & 63.3 & \color{blue}{+9.4} & \rev{64.2} & 59.8 & 68.1 & 66.4 & \color{blue}{+1.7} \\
\addlinespace[3pt]
Claude
  & Haiku 4.5 & \rev{68.6} & 70.4 & 72.0 & 70.9 & \textbf{\color{blue}{+1.1}} & \rev{55.8} & 58.5 & 60.0 & 64.8 & \color{red}{$-$4.8} \\
  & Sonnet 4.6 & \rev{68.6} & 72.8 & 74.1 & 63.3 & \color{blue}{+10.8} & \rev{69.5} & 72.3 & 71.1 & 63.2 & \color{blue}{+7.9} \\
  & Opus 4.6 & \rev{68.4} & 82.0 & 77.6 & 70.9 & \color{blue}{+6.7} & \rev{71.7} & 82.5 & 80.0 & 68.8 & \color{blue}{+11.2} \\
\addlinespace[3pt]
Gemini
  & 2.5 Flash & \rev{75.3} & 68.2 & 69.9 & 65.2 & \color{blue}{+4.7} & \rev{65.5} & 62.8 & 61.5 & 67.2 & \color{red}{$-$5.7} \\
  & 3 Flash & \textbf{82.6} & \textbf{87.5} & \textbf{83.9} & \textbf{80.4} & \color{blue}{+3.5} & \rev{\textbf{83.0}} & \textbf{90.9} & \textbf{91.1} & \textbf{87.2} & \color{blue}{+3.9} \\
\addlinespace[3pt]
Grok
  & 3-mini & \rev{81.1} & 62.6 & 64.3 & 59.5 & \color{blue}{+4.8} & \rev{64.4} & 67.8 & 71.1 & 69.6 & \color{blue}{+1.5} \\
  & 3 & \rev{75.7} & 63.2 & 60.8 & 66.5 & \color{red}{$-$5.6} & \rev{66.6} & 69.0 & 67.4 & 68.8 & \color{red}{$-$1.4} \\
  & 4-1 Fast & \rev{77.1} & 77.3 & 76.2 & 66.5 & \color{blue}{+9.8} & \rev{75.5} & 65.6 & 74.1 & 73.6 & \color{blue}{+0.5} \\
\addlinespace[2pt]
\rowcolor{gray!12}
\multicolumn{2}{l}{\textbf{Closed-source mean}}
  & \rev{\textbf{72.5}} & \textbf{66.8} & \textbf{67.5} & \textbf{63.5} & \textbf{\color{blue}{+4.0}}
  & \rev{\textbf{64.1}} & \textbf{65.0} & \textbf{66.9} & \textbf{66.5} & \textbf{\color{blue}{+0.4}} \\
\midrule
\multicolumn{12}{@{}l}{\textit{Open-Source Models}} \\[2pt]
Llama
  & 3.1-8B & \rev{51.1} & 45.7 & 48.3 & 46.2 & \color{blue}{+2.0} & \rev{41.8} & 47.7 & 51.9 & 44.8 & \color{blue}{+7.1} \\
  & 3.2-1B & \rev{60.0} & 54.9 & 56.6 & 53.8 & \color{blue}{+2.8} & \rev{50.7} & 60.8 & 52.6 & 49.6 & \color{blue}{+3.0} \\
  & 3.2-3B & \rev{60.7} & 45.7 & 54.5 & 42.4 & \color{blue}{+12.1} & \rev{49.3} & 50.5 & 57.0 & 44.8 & \color{blue}{+12.2} \\
  & 3.3-70B & \rev{75.5} & 54.4 & 55.2 & \textbf{67.7} & \color{red}{$-$12.5} & \rev{58.2} & 61.2 & 62.2 & \textbf{68.0} & \color{red}{$-$5.8} \\
\addlinespace[3pt]
Qwen
  & 3-8B & \rev{75.5} & 64.5 & 62.2 & 61.4 & \textbf{\color{blue}{+0.8}} & \rev{55.3} & \textbf{61.6} & \textbf{65.2} & 64.0 & \textbf{\color{blue}{+1.2}} \\
  & 3-14B & \rev{67.7} & 53.1 & 55.9 & 51.3 & \color{blue}{+4.7} & \rev{51.2} & 47.3 & 48.1 & 66.4 & \color{red}{$-$18.3} \\
  & 3-32B & \rev{72.2} & \textbf{69.1} & \textbf{67.1} & 64.6 & \color{blue}{+2.6} & \rev{57.1} & 50.1 & 49.6 & 61.6 & \color{red}{$-$12.0} \\
\addlinespace[2pt]
\rowcolor{gray!12}
\multicolumn{2}{l}{\textbf{Open-source mean}}
  & \rev{\textbf{66.1}} & \textbf{55.3} & \textbf{57.1} & \textbf{55.3} & \textbf{\color{blue}{+1.8}}
  & \rev{\textbf{51.9}} & \textbf{54.2} & \textbf{55.2} & \textbf{57.0} & \textbf{\color{red}{$-$1.8}} \\
\bottomrule
\end{tabular} 
\end{adjustbox}

\cutcaptionup
\caption{Accuracy (\%) under different in-context example conditions.
\rev{The \textsc{None} columns report the corrected-878 baseline; the example
columns preserve the original matched experiment. $\Delta_{\text{example}}$
is \textsc{Intervention-Ex} minus \textsc{Market-Ex} accuracy.}}
\label{tab:steering}
\end{table*}

\mypara{\rev{Example effects are stronger on intervention-truth targets}}
\rev{On intervention-truth targets, the mean $\Delta_{\text{example}}$ is
$+4.0$~pp for closed-source models and $+1.8$~pp for open-source models.
The largest individual gaps include Sonnet~4.6 ($+10.8$~pp) and GPT-5-mini
($+10.2$~pp). On market-truth targets, the corresponding means narrow to
$+0.4$~pp and $-1.8$~pp. The model-specific variation is substantial: Opus~4.6
has a gap of $+11.2$~pp, whereas Qwen3--14B has a gap of $-18.3$~pp.}

\mypara{\rev{Congruent examples do not reliably correct predictions}}
This interpretation is further supported by the fact that the gap is not driven by clean positive steering from a single example. If in-context examples provided effective ideological steering, congruent examples---\textsc{Intervention-Ex} on intervention-truth targets and \textsc{Market-Ex} on market-truth targets---should improve performance relative to the \textsc{None} baseline. Instead, the observed answer shifts are weak, uneven, and often fail to move in the corrective direction (Table~\ref{tab:icl_transition} in Appendix~\ref{app:icl_steering}).
\mypara{\rev{One-shot examples do not remove the main asymmetry}}
\rev{Taken together, these results show small group-level shifts and substantial
model-level heterogeneity. A single in-context example is insufficient to
reliably correct the ideological asymmetry documented in the main results,
although the behavioral evidence alone does not identify its internal cause.
Complementary GPT-4o confidence and paired open-model logit-lens diagnostics
appear in Figure~\ref{fig:confidence_diff} of
Appendix~\ref{subsec:icl_calibration} and Table~\ref{tab:logit_lens} of
Appendix~\ref{subsec:logit_lens}, respectively.}

\section{Discussion}
\label{sec:discussion}
\cutsectionup
\subsection{Implications for LLM-Assisted Economic Analysis}
\cutsubsectionup

Our results show that LLMs are not merely less accurate on ideologically contested economic causal questions; they are systematically more reliable in one ideological direction than the other. Across \rev{17} of 20 models, accuracy is higher when the empirical sign aligns with intervention-oriented expectations than with market-oriented expectations, with average gaps of $\rev{+8.5}$pp to $\rev{+14.2}$pp.  
 This makes the central concern not low accuracy alone, but directionally asymmetric reliability. Models can appear broadly competent while being substantially less dependable on one side of a contested economic question.

This asymmetry is qualitatively distinct from the opinion-level biases typically measured by political-compass-style instruments \citep{rozado2024, hartmann2023, santurkar2023, faulborn2025}, which are designed to elicit expressed preferences rather than detect systematic error against verifiable ground truth. This distinction matters because reasoning bias is harder for users to detect: when a model predicts that a minimum wage increase sustains employment, users are likely to interpret the answer as an empirical summary rather than as the product of a skewed prior. Previous work also suggests that users may not spontaneously discount ideological content in LLM outputs \citep{potter2024}. As a result, asymmetric reasoning errors may not only be difficult to detect in practice, but may also subtly influence downstream beliefs, judgments, and policy attitudes.

These findings imply that aggregate accuracy alone is an insufficient metric for evaluating LLMs in economic analysis. A model can achieve reasonable average performance while masking large directional disparities within ideologically contested cases. Our proposed metrics—the accuracy gap ($\Delta_{\text{acc}}$) and directional bias score ($B_{\text{dir}}$)—therefore provide a minimal but useful diagnostic for identifying failures that standard benchmarking would otherwise obscure. For LLM-assisted policy analysis, economic reporting, and decision support, such direction-aware evaluation should be treated as a basic robustness check rather than an optional extension.

\cutsubsectionup
\subsection{Possible Sources of the Bias}
\cutsubsectionup

\label{subsec:dis_bias_time_series}

\mypara{It is not just a prompting artifact} Several results suggest that the observed bias is unlikely to be a shallow prompting artifact. The intervention-truth advantage remains after difficulty matching, indicating that the pattern is not simply driven by market-truth items being intrinsically harder. One-shot in-context steering also fails to reliably reverse the bias: although examples can shift answer tendencies in some cases, they do not consistently eliminate the intervention-truth advantage. Taken together, these findings point to a more persistent directional tendency in model behavior rather than a transient prompt-level effect.

\cutparagraphup
\mypara{Part of this directional tendency may originate earlier in the training pipeline} A potential source of this directional tendency may lie in pretraining, though subsequent alignment processes can further amplify it. The base-versus-instruct comparison supports this interpretation. As shown in Table~\ref{tab:base_instruct}, a noticeable intervention-truth advantage is already present at the base-model stage, before instruction tuning in both Qwen 3-14B and Qwen 3-8B. Instruction tuning improves overall accuracy, but it does not remove this asymmetry; \rev{the accuracy delta widens from +11.5 to +16.4~pp for Qwen~3-14B and from +9.9 to +20.3~pp for Qwen~3-8B, while $B_{\text{dir}}$ shifts from $-2.0$ to +9.6 and from +1.9 to +16.9, respectively.} This pattern suggests that alignment does not simply wash out preexisting directional tendencies and may, in some cases, sharpen them, consistent with prior work showing that biases in pretraining corpora can propagate into downstream predictions \citep{feng2023}.

\begin{table}[tb]
\centering
\small
\setlength{\tabcolsep}{3.5pt}
\renewcommand{\arraystretch}{0.98}
\begin{tabular}{@{}llcccc@{}}
\toprule
Model & Variant & Acc\textsubscript{div} & Acc\textsubscript{neu} & $\Delta_{\text{acc}}$ & $B_{\text{dir}}$ \\
\midrule
Qwen 3-14B & Base     & 44.3 & 53.7 & \rev{+11.5} & \rev{$-$2.0} \\
           & Instruct & 52.2 & 69.3 & \rev{+16.4} & \rev{+9.6} \\
           & $\Delta$ & +7.9 & +15.6 & \rev{+4.9} & \rev{+11.5} \\
\midrule
Qwen 3-8B  & Base     & 46.3 & 55.7 & \rev{+9.9} & \rev{+1.9} \\
           & Instruct & 57.5 & 71.6 & \rev{+20.3} & \rev{+16.9} \\
           & $\Delta$ & +11.2 & +15.9 & \rev{+10.4} & \rev{+15.0} \\
\bottomrule
\end{tabular}

\cutcaptionup

\caption{\rev{Base vs.\ instruct comparison for Qwen pairs. Acc\textsubscript{div} and Acc\textsubscript{neu} denote accuracy on ideology-contested and non-contested items, respectively. $\Delta_{\text{acc}}$ is intervention-truth accuracy minus market-truth accuracy, and $B_{\text{dir}}$ is directional error bias.}}
\label{tab:base_instruct}

\end{table}

\begin{figure}[t!]
    \centering
    \includegraphics[width=0.97\linewidth]{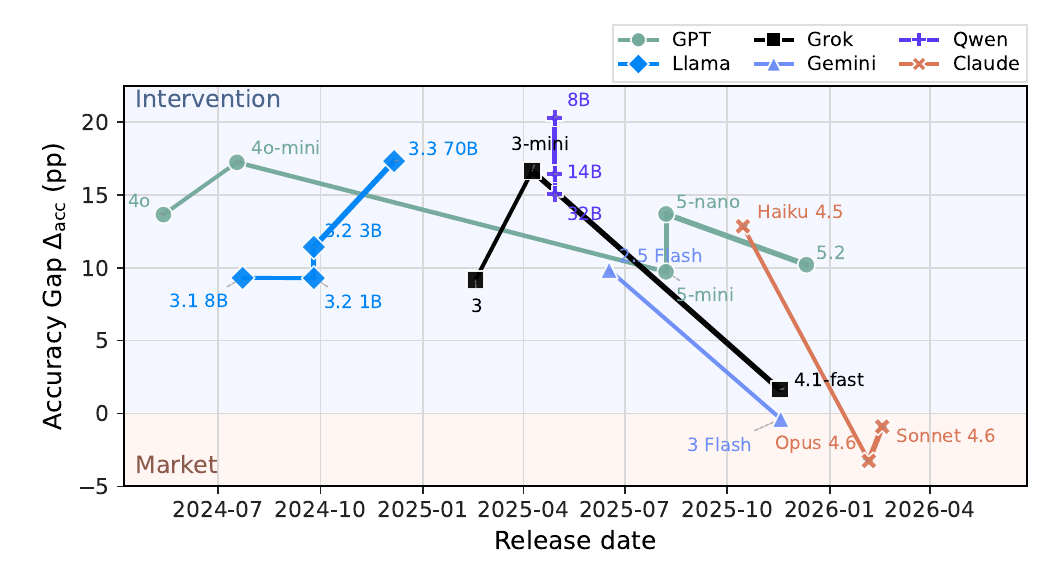}
    \cutcaptionup
    \caption{\rev{Accuracy gap ($\Delta_{\text{acc}}$)} by model release date.}
    \label{fig:trend}
\end{figure}

\cutparagraphup
\mypara{Post-training pipelines may matter as well} The family-level patterns in Figures~\ref{fig:main_results} and~\ref{fig:trend} suggest that the asymmetry cannot be reduced to a single uniform tendency. \rev{Claude Sonnet~4.6, Claude Opus~4.6, and Gemini~3 Flash show small market-truth advantages,} while within the GPT and Grok families the magnitude of the gap varies non-monotonically, only partially tracking scale-based explanations~\citep{exler2025}. These patterns suggest that family-specific post-training pipelines may amplify or suppress preexisting directional tendencies in different ways. Moreover, Figure~\ref{fig:trend} shows that this pattern is not static: earlier models cluster toward the intervention side, while the most recent releases trend toward market-truth accuracy, suggesting the bias is evolving alongside model generations. Since the alignment details of commercial models remain proprietary, however, disentangling these contributions remains an important direction for future work.
\cutparagraphup
\mypara{The distribution of ground-truth signs in the benchmark may also contribute to the bias} In the ideology-contested evaluation subset, the ground-truth sign in each triplet is anchored in the context-specific causal estimate supported by the paper’s rigorously designed empirical strategy. This is precisely why we focus on top-tier economics and finance journals: the goal is to evaluate model behavior against empirical claims that have survived unusually rigorous peer review and robustness scrutiny, not against arbitrary reported estimates. At the same time, rigor at the level of individual studies does not imply that the benchmarked distribution of signs perfectly matches the latent population distribution of true causal effects. Even when individual papers are carefully designed and peer-reviewed, the published literature may still reflect selection in which questions are studied, which results become visible and publishable, and which context-specific findings accumulate in the record. In ideologically contested domains, such selection and heterogeneity may themselves generate distributional skew in the benchmark. \rev{Consistent with this concern, after controlling for model, theme, and difficulty, adding empirical-sign fixed effects attenuates the intervention-truth coefficient by 10.7\% (from 0.455 to 0.406); the fully adjusted estimate remains positive and significant ($p<0.001$; Appendix~\ref{app:difficulty}).}

Taken together, these results do not imply that the observed bias reflects ideological preference alone. A more cautious interpretation is that the pattern arises from an interaction among pretraining, alignment, family-specific post-training pipelines, and distributional regularities inherited from the empirical and discursive environments on which models are trained. Regardless of its precise source, however, the directional bias we document in LLM economic causal reasoning is itself policy-relevant.

\cutsubsectionup
\subsection{Future Directions}
\cutsubsectionup
A key next step is to distinguish more clearly between memory-based retrieval and mechanism-based reasoning. When models answer ideologically contested economic causal items correctly, they may do so either by recalling paper-specific empirical claims or by reasoning through competing economic mechanisms from the provided context. Disentangling these pathways would help clarify whether the bias we document primarily reflects memorized regularities in the literature, asymmetries in mechanism selection, or an interaction between the two.

\section*{\rev{Acknowledgments}}
\rev{This work was supported by the National Research Foundation of Korea (NRF) grants funded by the Korean government (MSIT) (Grant Nos. RS-2022-NR068758 and RS-2025-00563196). This work was also supported by the AI Science Hub Program of the Ministry of Science and ICT (Grant No. N10260126).}

\section*{\rev{LLM Usage Disclosure}}
\rev{Large language models were used both as objects of evaluation and as components of the research pipeline. As described in the paper, LLMs generated ideology-conditioned annotations and causal-sign predictions, and open-weight models were used for the internal-representation analysis. OpenAI Codex assisted with camera-ready prose editing, LaTeX and figure code, bibliography checking, and cross-file consistency checks. Model outputs did not replace EconCausal's paper-specific empirical ground-truth labels. The authors reviewed all model-assisted outputs and revisions and take full responsibility for the final manuscript.}

\newpage

\bibliography{colm2026_conference}
\bibliographystyle{colm2026_conference}
\appendix
\section*{\rev{Appendix}}
\section{Economic Subfield Taxonomy}
\label{app:subfield}

Ideology-contested causal triplets are grouped into seven economic subfields based on Journal of Economic Literature (JEL) classification codes. Table~\ref{tab:subfield_mapping} presents the mapping rules and sample sizes.

\begin{table}[h]
\centering
\small
\setlength{\tabcolsep}{4pt}
\caption{JEL code mapping to economic subfields.}
\label{tab:subfield_mapping}
\begin{tabular}{llr p{5.8cm}}
\toprule
\textbf{Subfield} & \textbf{JEL Prefix} & \rev{\textbf{\#Memberships}} & \textbf{Definition of JEL code(s)} \\
\midrule
Taxation               & \texttt{H2*}                & \rev{77}  & Taxation, Subsidies, and Revenue \\
Healthcare             & \texttt{I1*}                & \rev{101} & Health \\
Education              & \texttt{I2*}                & \rev{60}  & Education and Research Institutions \\
Welfare/Redistribution & \texttt{H5*}, \texttt{I3*}  & \rev{92}  & National Government Expenditures and Related Policies; Welfare, Well-Being, and Poverty \\
Labor                  & \texttt{J*}                 & \rev{218} & Labor and Demographic Economics \\
Financial Regulation   & \texttt{G*}                 & \rev{220} & Financial Economics \\
Trade                  & \texttt{F1*}                & \rev{38}  & Trade \\

Other                  & ---                         & \rev{201} & Remaining codes (excluded) \\
\bottomrule
\end{tabular}
\end{table}

\rev{A triplet may be associated with multiple JEL codes. Each code contributes
one vote to its mapped subfield, and only the subfield(s) attaining the largest
vote count receive the item's analysis weight. A unique winner receives weight
1; exact ties split weight equally across the tied subfields. For example, an
item with one \texttt{G35} and one \texttt{H25} code contributes weight 0.5 to
financial regulation and 0.5 to taxation. Table~\ref{tab:subfield_mapping}
reports unweighted top-vote membership counts (so a tied item is counted in
each tied subfield), whereas Figure~\ref{fig:subfield_gap} uses the fractional
analysis weights. Codes outside the seven focal categories are labeled
\emph{other} and excluded from the plotted subfield comparison.}

\begin{figure}[h]
    \centering
    \includegraphics[width=0.7\linewidth]{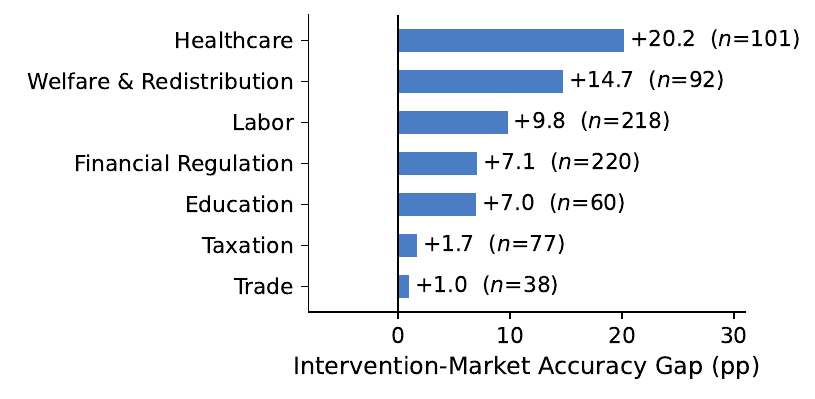}
    \vspace{-3mm}
    \caption{\rev{Vote-weighted mean accuracy gap of 20 models across economic subfields.}}
    \label{fig:subfield_gap}
\end{figure}%
\cutsubsectionup

\section{Full Model-Level Results on Ideology-contested Items}
\label{app:full_model}

\begin{table*}[h]

\centering
\small
\setlength{\tabcolsep}{5pt}
\renewcommand{\arraystretch}{1.05}
\resizebox{\textwidth}{!}{%
\begin{tabular}{@{}ll cccc cc@{}}
\toprule
& & \multicolumn{4}{c}{\textit{Accuracy (\%)}} & \multicolumn{2}{c}{\textit{Bias}} \\
\cmidrule(lr){3-6} \cmidrule(lr){7-8}
\textbf{Family} & \textbf{Model}
  & Non-contested & Contested
  & Intervention & Market
  & Acc. Gap & \rev{Error Direction Bias (\%)} \\
\midrule
\multicolumn{8}{@{}l}{\textit{Closed-Source Models}} \\[2pt]
OpenAI
  & GPT-4o-mini          & 63.2 & \rev{49.3} & \rev{65.5} & \rev{48.2} & \rev{+17.2} & \rev{+6.3} \\
  & GPT-4o               & 69.4 & \rev{57.1} & \rev{71.6} & \rev{58.0} & \rev{+13.6} & \rev{+6.0} \\
  & GPT-5-nano           & 70.1 & \rev{53.5} & \rev{67.1} & \rev{53.4} & \rev{+13.7} & \rev{+7.4} \\
  & GPT-5-mini           & 73.1 & \rev{55.4} & \rev{66.9} & \rev{57.1} & \rev{+9.7}  & \rev{+3.7} \\
  & GPT-5.2              & 75.8 & \rev{62.9} & \rev{74.4} & \rev{64.2} & \rev{+10.2} & \rev{+0.8} \\
\addlinespace[3pt]
Claude
  & Haiku 4.5            & 74.2 & \rev{55.7} & \rev{68.6} & \rev{55.8} & \rev{+12.8} & \rev{+6.2} \\
  & Sonnet 4.6           & 76.3 & \rev{63.1} & \rev{68.6} & \rev{69.5} & \rev{$-$0.9}  & \rev{$-$5.9} \\
  & Opus 4.6             & 77.7 & \rev{66.2} & \rev{68.4} & \rev{71.7} & \rev{$-$3.3}  & \rev{$-$10.2} \\
\addlinespace[3pt]
Gemini
  & 2.5 Flash            & 75.9 & \rev{61.1} & \rev{75.3} & \rev{65.5} & \rev{+9.8} & \rev{+10.7} \\
  & 3 Flash              & \textbf{83.5} & \rev{\textbf{76.3}} & \textbf{82.6} & \rev{\textbf{83.0}} & \rev{$-$0.4}  & \rev{$-$7.3} \\
\addlinespace[3pt]
Grok
  & 3-mini               & 76.7 & \rev{63.4} & \rev{81.1} & \rev{64.4} & \rev{+16.6} & \rev{+14.5} \\
  & 3                    & 74.5 & 61.2 & \rev{75.7} & \rev{66.6} & \rev{+9.2} & \rev{+1.6} \\
  & 4-1 Fast             & 78.2 & \rev{65.8} & \rev{77.1} & \rev{75.5} & \rev{+1.6}  & \rev{$-$3.9} \\
\addlinespace[2pt]
\rowcolor{gray!12}
\multicolumn{2}{l}{\textbf{Closed-source mean}} 
  & \textbf{74.5} & \rev{\textbf{60.8}} & \rev{\textbf{72.5}} & \rev{\textbf{64.1}}
  & \rev{\textbf{+8.5}} & \rev{\textbf{+2.3}} \\
\midrule
\multicolumn{8}{@{}l}{\textit{Open-Source Models}} \\[2pt]
Llama
  & 3.1-8B               & 49.3 & \rev{42.0} & \rev{51.1} & \rev{41.8} & \rev{+9.3}  & \rev{$-$3.9} \\
  & 3.2-1B               & 56.4 & \rev{48.1} & \rev{60.0} & \rev{50.7} & \rev{+9.3} & \rev{+0.8} \\
  & 3.2-3B               & 56.7 & \rev{48.4} & \rev{60.7} & \rev{49.3} & \rev{+11.4} & \rev{+5.2} \\
  & 3.3-70B              & \textbf{72.1} & \rev{57.7} & \rev{75.5} & \rev{58.2} & \rev{+17.3} & \rev{+12.5} \\
\addlinespace[3pt]
Qwen
  & 3-8B                 & 71.6 & \rev{57.5} & \rev{75.5} & \rev{55.3} & \rev{+20.3} & \rev{+16.9} \\
  & 3-14B                & 69.3 & \rev{52.2} & \rev{67.7} & \rev{51.2} & \rev{+16.4} & \rev{+9.6} \\
  & 3-32B                & 71.1 & \rev{56.2} & \rev{72.2} & \rev{57.1} & \rev{+15.0} & \rev{+10.0} \\
\addlinespace[2pt]
\rowcolor{gray!12}
\multicolumn{2}{l}{\textbf{Open-source mean}} 
  & \textbf{63.8} & \rev{\textbf{51.7}} & \rev{\textbf{66.1}} & \rev{\textbf{51.9}}
  & \rev{\textbf{+14.2}} & \rev{\textbf{+7.3}} \\
\bottomrule
\end{tabular}}
\vspace{4pt}
\caption{Full model-level performance and ideological bias. \rev{\textit{Contested} accuracy uses all 1,056 ideology-contested items; intervention-truth and market-truth accuracy, Acc.\ Gap, and Error Direction Bias use the 507/371 directionally aligned subsets.} Positive \textit{Acc.\ Gap} indicates higher accuracy on intervention-truth items than on market-truth items, and positive \textit{Error Direction Bias} indicates intervention-leaning errors. \rev{Acc.\ Gap is reported in percentage points, and Error Direction Bias is reported in percent.}}
\label{tab:main_results}

\end{table*}

\rev{Table~\ref{tab:main_results} reports each model's accuracies and the two
directional metrics. The accuracy gap is
$\Delta_{\mathrm{acc}}=\mathrm{Acc}_{\mathrm{intervention}}-
\mathrm{Acc}_{\mathrm{market}}$; $B_{\mathrm{dir}}$ measures whether errors
lean toward intervention- or market-oriented expectations. The overall
contested column uses all 1,056 items, whereas the directional columns and
metrics use 507 intervention-truth and 371 market-truth items.}

\section{Difficulty Scoring, Matching, and Regression}
\label{app:difficulty}

\mypara{Difficulty scoring}
Each causal triplet in the \rev{878-item directional subset} was scored on a 1--5 difficulty scale by GPT-5-mini.
The prompt instructed the model to evaluate five dimensions:
\begin{itemize}[nosep,leftmargin=1.5em]
  \item \textbf{Domain knowledge}: depth of economics expertise required;
  \item \textbf{Context dependence}: reliance on the provided socio-economic context;
  \item \textbf{Ambiguity / confoundability}: plausibility of competing causal directions;
  \item \textbf{Causal reasoning complexity}: number of intermediate steps or mechanisms;
  \item \textbf{Evidence sufficiency}: whether the context provides enough information to determine the sign.
\end{itemize}
A composite \emph{overall difficulty} score (1--5) was then derived from these five dimensions.
The full scoring prompt is shown below.

\begin{figure}[h]
\begin{tcolorbox}[
  colback=black!0!white, colframe=black!20!white,
  colbacktitle=black!10!white, coltitle=blue!20!black,
  title={\small\textbf{Difficulty Scoring Prompt}}
]
\small
\setlength{\parindent}{0pt}
\setlength{\parskip}{4pt}

\textbf{System Prompt:}

You are an expert economist. Given an economic causal triplet (treatment $\rightarrow$ outcome with a stated sign), rate how difficult it is to correctly identify the causal sign. Use these 5 dimensions (each 1--5) plus an overall score.

\begin{enumerate}[nosep,leftmargin=1.5em]
  \item \textbf{Domain Knowledge} (1--5): 1 = common sense sufficient, 3 = intermediate policy/market knowledge needed, 5 = requires specialized theory or identification strategies.
  \item \textbf{Context Dependence} (1--5): 1 = sign is stable regardless of setting, 3 = context matters moderately, 5 = sign is entirely pinned down by specific institutional/temporal/population conditions.
  \item \textbf{Ambiguity / Confoundability} (1--5): 1 = sign is unambiguous, 3 = moderately plausible alternative, 5 = opposite sign nearly as plausible without empirical analysis.
  \item \textbf{Causal Reasoning Complexity} (1--5): 1 = direct one-step link, 3 = multiple mediating channels, 5 = requires reasoning through GE feedbacks, selection, and composition effects simultaneously.
  \item \textbf{Evidence Sufficiency} (1--5): 1 = context clearly determines the sign, 3 = reasonable inference possible despite gaps, 5 = information so incomplete that None/Mixed is a serious contender.
  \item \textbf{Overall Difficulty} (1--5): Your holistic judgment of how hard this triplet is, considering all dimensions above.
\end{enumerate}

Evaluate PURELY based on economic reasoning. Do NOT let political implications influence your ratings.

Respond in JSON only, no other text:

{\ttfamily\small
\{"domain\_knowledge":<int>, "context\_dependence":<int>, "ambiguity":<int>, "causal\_complexity":<int>, "evidence\_sufficiency":<int>, "overall\_difficulty":<int>, "justification":"<1--2 sentences>"\}}

\vspace{6pt}
\textbf{User Prompt Template:}

Rate the difficulty of this economic causal triplet.

{\ttfamily
Treatment: \{treatment\} \\
Outcome: \{outcome\} \\
Sign: \{sign\} \\
Context: \{context\}}

\end{tcolorbox}
\caption{Prompt used for difficulty scoring (GPT-5-mini).}
\label{fig:difficulty_prompt}
\end{figure}

\mypara{Matching procedure}
We match intervention-truth and market-truth triplets one-to-one within each
\emph{theme $\times$ difficulty} cell. For JEL-based policy theme $t$ and difficulty level $d$,
we compute $m_{t,d} = \min(n_{t,d}^{\mathrm{intervention}},\, n_{t,d}^{\mathrm{market}})$ and draw
$m_{t,d}$ triplets per side using the fixed random seed 42.
Cells where either side has zero count are excluded, and all 20 models' prediction rows on the selected triplets are retained for re-evaluation.
This procedure yielded \rev{352} triplets per side (\rev{704} total).

\mypara{Logistic regression}
\rev{As a complementary check, we estimate a logistic regression on the full
878-item directional subset rather than restricting the regression to the
704-item matched sample:}
\begin{equation}
  \mathrm{logit}\bigl(P(\textit{correct}_{i,m})\bigr)
    = \beta_0
    + \beta_1\,\mathbb{1}[\textit{ground\_truth} = \textit{intervention}]_i
    + \boldsymbol{\gamma}\,\mathbf{D}_i
    + \boldsymbol{\delta}\,\mathbf{T}_i
    + \boldsymbol{\eta}\,\mathbf{M}_m
    + \boldsymbol{\rho}\,\mathbf{S}_i
\end{equation}
where $\mathbf{D}_i$ denotes difficulty-level dummies, $\mathbf{T}_i$ theme fixed effects,
and $\mathbf{M}_m$ model fixed effects. \rev{$\mathbf{S}_i$ denotes empirical
ground-truth-sign fixed effects.} The coefficient of interest is $\beta_1$, which
captures the residual accuracy advantage of intervention-truth items after controlling for
difficulty, subfield, model heterogeneity, \rev{and empirical sign}.

\rev{We retain all 17,560 model--item observations and cluster standard errors
by triplet. Table~\ref{tab:full878_regression} adds each set of controls
sequentially; the final row corresponds to the specification above.}

\begin{table}[h!]
\centering
\small
{\color{revisionpurple}
\begin{tabular}{lrrrr}
\toprule
Specification & $\hat{\beta}_1$ & Clustered SE & $z$ & $p$ \\
\midrule
Bivariate & 0.463 & 0.079 & 5.84 & $<0.001$ \\
$+$ model FE & 0.476 & 0.082 & 5.83 & $<0.001$ \\
$+$ theme FE & 0.470 & 0.083 & 5.68 & $<0.001$ \\
$+$ difficulty FE & 0.455 & 0.082 & 5.56 & $<0.001$ \\
$+$ empirical-sign FE & 0.406 & 0.077 & 5.28 & $<0.001$ \\
\bottomrule
\end{tabular}}
\cutcaptionup
\caption{\rev{Full-878 logistic regressions of item-level correctness with
sequentially added controls. Each row retains the controls in the preceding row.}}
\label{tab:full878_regression}
\end{table}

\rev{After controlling for model, theme, and difficulty, adding empirical-sign
fixed effects attenuates $\hat{\beta}_1$ by 10.7\% (from 0.455 to 0.406). The
fully adjusted estimate remains positive and statistically significant.}

\section{ICL Example Details}
\label{app:icl_details}

\subsection{Example Matching}
\label{app:icl_matching}

For each ideology-contested target triplet, we define four prompting conditions:
\textsc{None}, \textsc{Non-contested}, \textsc{Intervention-Ex}, and \textsc{Market-Ex}.
A candidate example is matched to a target if it satisfies the following conditions:
\begin{itemize}
    \item the source papers of the target and example share at least two JEL codes; and
    \item their Jaccard similarity over JEL codes is at least $0.5$.
\end{itemize}
Let $W_t$ and $W_e$ denote the sets of JEL codes assigned to the source papers from which the target and example triplets are extracted. We compute similarity as
\[
J(W_t, W_e) = \frac{|W_t \cap W_e|}{|W_t \cup W_e|},
\qquad
|W_t \cap W_e| \geq 2,
\qquad
J(W_t, W_e) \geq 0.5.
\]
Table~\ref{tab:icl_matching_counts} reports the number of available instances
in each condition. The paired \textsc{Intervention-Ex}/\textsc{Market-Ex}
conditions contain 125--158 matched instances per model.

\begin{table}[h]
\centering
\small
\begin{tabular}{lcc}
\toprule
\textbf{Condition} & \textbf{Intervention-truth} & \textbf{Market-truth} \\
\midrule
\textsc{None} & \rev{507} & \rev{371} \\
\textsc{Non-contested} & 1325 & 761 \\
\textsc{Intervention-Ex} & 143 & 135 \\
\textsc{Market-Ex} & 158 & 125 \\
\bottomrule
\end{tabular}
\caption{Available instances per ICL condition and orientation.
\rev{\textsc{None} reports the corrected-878 reference; the other rows preserve
the original matched counts.}}
\label{tab:icl_matching_counts}
\end{table}

\subsection{Steering Effects}
\label{app:icl_steering}

Table~\ref{tab:icl_transition} reports the mean answer shifts across 20 models when a single in-context example is introduced, relative to the \textsc{None} baseline. Congruent examples---\textsc{Intervention-Ex} on intervention-truth targets and \textsc{Market-Ex} on market-truth targets---do not yield consistent corrective shifts, indicating that a single example is insufficient to reliably steer model predictions.

\begin{table}[h]
\centering
\small
\setlength{\tabcolsep}{5pt}
\renewcommand{\arraystretch}{1.0}
\begin{tabular}{@{}lcc@{}}
\toprule
 & \textbf{Intervention-truth} & \textbf{Market-truth} \\
\midrule
\textsc{Intervention-Ex}: $\Delta$ intervention-aligned rate & $-2.8$ & $-0.8$ \\
\textsc{Market-Ex}: $\Delta$ market-aligned rate      & $+3.1$ & $-4.6$ \\
\bottomrule
\end{tabular}
\caption{Answer shifts from \textsc{None} (20-model mean, pp). Congruent
conditions (top-left and bottom-right cells) show weak or negative shifts.}
\label{tab:icl_transition}
\end{table}

\subsection{\rev{GPT-4o Confidence Diagnostic}}
\label{subsec:icl_calibration}

{\color{revisionpurple}
We examine whether confidence helps explain how intervention- and market-aligned
examples produce different behavioral effects across target types. We analyze
GPT-4o because it is the strongest evaluated model whose API provided stable
token-level log-probabilities, and define confidence as the exponentiated
log-probability of the emitted sign-bearing token. Because the two example
conditions contain different matched targets, the comparisons below are
descriptive.
\begin{itemize}[leftmargin=1.5em,itemsep=2pt,topsep=2pt]
    \item \textbf{Intervention-truth targets:} \textsc{Market-Ex} yields higher
    confidence than \textsc{Intervention-Ex} (0.711 vs.\ 0.642;
    $\Delta\mu=+0.068$, $KS=0.159$, $p=0.039$), but slightly lower accuracy
    (43.7\% vs.\ 45.5\%). This high-confidence, lower-accuracy combination is
    consistent with a backfire pattern in which a contrastive example may
    increase decisional certainty without correcting the judgment.
    \item \textbf{Market-truth targets:} The confidence distributions are not
    significantly different (0.750 vs.\ 0.718; $\Delta\mu=+0.032$,
    $KS=0.090$, $p=0.625$), while \textsc{Market-Ex} improves accuracy by
    10.6~pp (63.2\% vs.\ 52.6\%). This pattern is consistent with steering
    acting more through answer orientation than confidence: the model can
    remain similarly confident while reaching different conclusions.
\end{itemize}
Together, the two patterns suggest target-dependent effects: contrastive
examples may raise certainty without improving correctness on intervention-truth
targets, whereas they may redirect answers without a comparable confidence
change on market-truth targets. These behavioral results do not establish the
underlying mechanism, so this interpretation remains tentative.
}

\begin{figure}[htbp]
    \centering
    \includegraphics[width=0.9\linewidth]{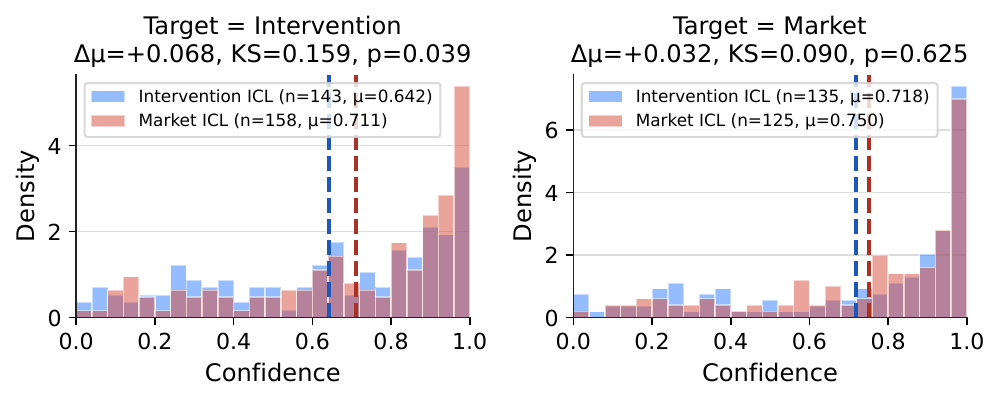}
    \cutcaptionup
    \caption{\rev{GPT-4o confidence under intervention- and market-aligned
    examples for intervention-truth (left) and market-truth (right) targets.}}
    \label{fig:confidence_diff}
\end{figure}

\subsection{\rev{Logit-Lens Diagnostic}}
\label{subsec:logit_lens}

\rev{Separately from the behavioral ICL analysis, we apply a logit lens to
Qwen3--14B, Qwen3--32B, and Llama~3.3--70B on a strict paired subset of 89
targets (49 intervention-truth and 40 market-truth). Prompts are truncated
immediately before the fixed sign-token pair. A two-way softmax defines
$m_{\mathrm{int}}=p(s_{\mathrm{int}})-p(s_{\mathrm{mkt}})$ and
$\Delta m=m_{\mathrm{int}}(\textsc{Intervention-Ex})-
m_{\mathrm{int}}(\textsc{Market-Ex})$. Table~\ref{tab:logit_lens} reports
final-layer target means and 95\% percentile intervals from 5,000
nonparametric target-level resamples.}

\begin{table}[h!]
\centering
\footnotesize
\setlength{\tabcolsep}{3pt}
{\color{revisionpurple}
\begin{tabular}{lcc}
\toprule
Model & Intervention-truth $\Delta m$ & Market-truth $\Delta m$ \\
\midrule
Qwen3--14B & $+0.087\;[-0.03,+0.21]$ & $-0.067\;[-0.26,+0.13]$ \\
Qwen3--32B & $+0.019\;[-0.06,+0.09]$ & $-0.029\;[-0.13,+0.06]$ \\
Llama~3.3--70B & $+0.088\;[-0.00,+0.20]$ & $+0.108\;[-0.01,+0.24]$ \\
\bottomrule
\end{tabular}}
\cutcaptionup
\caption{\rev{Mean example-induced orientation contrast in the paired
logit-lens analysis. Positive values indicate relative movement toward the
intervention-oriented sign; brackets report 95\% confidence intervals.}}
\label{tab:logit_lens}
\end{table}

\rev{All six intervals include zero. The largest positive contrasts are
Llama~3.3--70B on intervention-truth targets ($+0.088$, 95\% CI
$[-0.00,+0.20]$) and market-truth targets ($+0.108$, 95\% CI
$[-0.01,+0.24]$). Thus, consistent with the behavioral results, one example
does not produce a substantial or reliable change in sign orientation.}

\section{\rev{Prompt Robustness}}
\label{app:prompt_robustness}

\rev{We re-evaluate the current 878-item directional subset under the Baseline
(the original main prompt) and five prompt variants---Simple, Expert, Sign
Random Mix, Market First, and Intervention First---using three runs per
condition across 16 models. Gemini~2.5 Flash and the three Grok models in the
main 20-model panel are excluded from this rerun panel. Invalid or unparsable
predictions are excluded from the corresponding cell denominator.
Table~\ref{tab:prompt_ablation_corrected878} reports the corrected-878 results.}

\begin{table}[h!]
\centering
\small
\setlength{\tabcolsep}{5pt}
\begin{tabular}{lcccc}
\toprule
Condition & Closed $\Delta_{\mathrm{acc}}$ & Open $\Delta_{\mathrm{acc}}$ & Closed $B_{\mathrm{dir}}$ & Open $B_{\mathrm{dir}}$ \\
\midrule
\rev{Baseline} & \rev{$+8.2 \pm 0.5$} & \rev{$+12.9 \pm 1.3$} & \rev{$+0.9 \pm 0.8$} & \rev{$+4.5 \pm 2.4$} \\
Simple & \rev{$+9.6 \pm 0.2$} & \rev{$+14.6 \pm 0.5$} & \rev{$+2.0 \pm 0.1$} & \rev{$+6.3 \pm 0.9$} \\
Expert & \rev{$+7.4 \pm 0.7$} & \rev{$+13.8 \pm 0.3$} & \rev{$+0.4 \pm 0.3$} & \rev{$+5.3 \pm 0.5$} \\
Sign Random Mix & \rev{$+8.7 \pm 0.4$} & \rev{$+13.7 \pm 0.2$} & \rev{$+1.7 \pm 0.4$} & \rev{$+4.9 \pm 0.5$} \\
Market First & \rev{$+8.5 \pm 0.7$} & \rev{$+14.2 \pm 0.4$} & \rev{$+0.8 \pm 0.7$} & \rev{$+6.0 \pm 0.6$} \\
Intervention First & \rev{$+8.1 \pm 0.2$} & \rev{$+13.1 \pm 0.4$} & \rev{$+1.9 \pm 0.5$} & \rev{$+5.0 \pm 0.2$} \\
\bottomrule
\end{tabular}
\caption{Prompt-robustness results on the corrected 878-item directional subset. \rev{Baseline uses the original main prompt.} Values are means $\pm$ sample standard deviations over three runs; \rev{$\Delta_{\mathrm{acc}}$ is reported in percentage points and $B_{\mathrm{dir}}$ in percent.}}
\label{tab:prompt_ablation_corrected878}
\end{table}

\section{Prompts}
\label{app:prompts}

\begin{figure}[h]
\begin{tcolorbox}[
  colback=black!0!white, colframe=black!20!white,
  colbacktitle=black!10!white, coltitle=blue!20!black,
  title={\small\textbf{Main Result Sign Prediction Prompt}}
]
\small
\setlength{\parindent}{0pt}
\setlength{\parskip}{4pt}

\textbf{Role:}

You are an expert economist.

\textbf{Task:}

Given a context and a treatment--outcome pair, predict the most likely sign of the causal effect.

\textbf{Sign Definitions:}
\begin{itemize}[nosep,leftmargin=1.5em]
  \item \textbf{\texttt{+}}: The treatment increases the outcome (positive and statistically significant effect).
  \item \textbf{\texttt{-}}: The treatment decreases the outcome (negative and statistically significant effect).
  \item \textbf{\texttt{None}}: No statistically significant effect.
  \item \textbf{\texttt{mixed}}: The effect varies across subgroups or specifications.
\end{itemize}

\vspace{6pt}
\textbf{Prompt:}

{\ttfamily
\# Context: \{context\} \\
\\
\# Treatment--Outcome Pair \\
Treatment: \{treatment\} \\
Outcome: \{outcome\} \\
\\
\# Output Format \\
Respond with a JSON object containing: \\
- predicted\_sign: "+", "-", "None", or "mixed" \\
- reasoning: A concise explanation of your reasoning
}

\end{tcolorbox}
\caption{Prompt used for main-result sign prediction.}
\label{fig:main_result_prompt}
\end{figure}

\begin{figure}[h]
\begin{tcolorbox}[
  colback=black!0!white, colframe=black!20!white,
  colbacktitle=black!10!white, coltitle=blue!20!black,
  title={\small\textbf{Example-Augmented Sign Prediction Prompt}}
]
\small
\setlength{\parindent}{0pt}
\setlength{\parskip}{4pt}

\textbf{Role:}

You are an expert economist.

\textbf{Task:}

You are given examples related to the target treatment--outcome pair, potentially with different causal signs.
Predict the most likely sign for the treatment--outcome pair in the target context.

\textbf{Sign Definitions:}
\begin{itemize}[nosep,leftmargin=1.5em]
  \item \textbf{\texttt{+}}: The treatment increases the outcome (positive and statistically significant effect).
  \item \textbf{\texttt{-}}: The treatment decreases the outcome (negative and statistically significant effect).
  \item \textbf{\texttt{None}}: No statistically significant effect.
  \item \textbf{\texttt{mixed}}: The effect varies across subgroups or specifications.
\end{itemize}

\vspace{6pt}
\textbf{Prompt:}

{\ttfamily
\# Reference Examples: \{examples\} \\
\\
\# Target \\
Treatment: \{treatment\} \\
Outcome: \{outcome\} \\
Context: \{context\_new\} \\
\\
\# Output Format \\
Respond with a JSON object containing: \\
- predicted\_sign: "+", "-", "None", or "mixed" \\
- reasoning: A concise explanation of your reasoning
}

\end{tcolorbox}
\caption{Prompt used for example-augmented sign prediction.}
\label{fig:example_augmented_prompt}
\end{figure}

\setlist[itemize]{nosep,leftmargin=1.3em}
\setlist[enumerate]{nosep,leftmargin=1.3em}

\begin{figure}[t]
\begin{tcolorbox}[
  colback=black!0!white,
  colframe=black!20!white,
  colbacktitle=black!10!white,
  coltitle=blue!20!black,
  title={\small\textbf{Ideological Expected Sign Annotation Prompt}},
  boxsep=2pt,
  left=3pt,right=3pt,top=3pt,bottom=3pt
]
\footnotesize
\setlength{\parindent}{0pt}
\setlength{\parskip}{2pt}

\textbf{System Prompt:}

You are an annotation assistant for economics causal triplets. Your job is to read the given context, treatment, and outcome, and assign each ideology's expected causal sign.

\textbf{Rules:}
\begin{enumerate}
  \item Output MUST be valid JSON only. No extra text.
  \item Infer expected signs from ideological priors, NOT from the paper's empirical result.
  \item Most causal triplets are NOT ideologically contested. If both ideologies would expect the same sign, assign the same sign to both.
  \item Use \texttt{null} when no meaningful ideological prior exists for either side. Use the same non-null sign only when both sides would genuinely expect the same directional effect.
  \item Also provide one brief reasoning string of at most 100 words explaining the sign judgment.
\end{enumerate}

\textbf{Theoretical Framework} (based on Feldman \& Johnston, 2014; Alesina \& Giuliano, 2011)

Political ideology has at least two partially independent dimensions; here we focus ONLY on the \textbf{economic} dimension. The economic dimension captures attitudes toward the appropriate role of government versus the market in organizing economic life.

\begin{itemize}
  \item \textbf{economic\_conservative}: favors markets, limited government, lower taxes, deregulation, flexible labor markets, and skepticism toward redistribution; emphasizes incentives, efficiency, and moral hazard.
  \item \textbf{economic\_liberal}: favors active government intervention, redistribution, regulation, labor protections, public services, and social insurance; emphasizes inequality, unequal opportunity, structural disadvantage, and market failure.
\end{itemize}

\textbf{Sensitivity Criterion:}

A triplet is ideologically contested ONLY when disagreement over the expected sign stems directly from the government-intervention-vs-market-freedom divide.

Do NOT assign different signs merely because:
\begin{itemize}
  \item the topic is political or controversial,
  \item one ideology might normatively prefer the treatment or dislike the outcome,
  \item the setting involves firms, banks, or investors.
\end{itemize}

If no short, direct link from that divide to a different expected sign exists, assign \texttt{null} to both ideologies.

\textbf{Task:}

For each triplet, assign what each ideology would \emph{expect} as the causal sign of treatment $\rightarrow$ outcome. Judge only the direction of the causal link.

\textbf{Sign Options:}
\begin{itemize}
  \item \textbf{\texttt{+}}: positive causal effect
  \item \textbf{\texttt{-}}: negative causal effect
  \item \textbf{\texttt{None}}: no significant causal effect
  \item \textbf{\texttt{Mixed}}: mixed or context-dependent effects
  \item \textbf{\texttt{null}}: no ideological prior
\end{itemize}

\textbf{Examples where signs differ:}
\begin{itemize}
  \item minimum wage increase $\rightarrow$ employment
  \item corporate tax rate $\rightarrow$ business investment
\end{itemize}

\textbf{Examples where signs are the same:}
\begin{itemize}
  \item oil price shock $\rightarrow$ inflation
  \item R\&D spending $\rightarrow$ firm productivity
\end{itemize}

Respond in JSON only, no other text:

{\ttfamily\footnotesize
\{"ideological\_expected\_signs": \{"economic\_liberal\_expected\_sign":"<+|-|None|Mixed|null>", "economic\_conservative\_expected\_sign":"<+|-|None|Mixed|null>"\}, "reasoning":"<brief reasoning, <=100 words>"\}
}

\vspace{4pt}
\textbf{User Prompt Template:}

{\ttfamily\footnotesize
Paper Title: \{title\} \\
Context: \{context\} \\
Treatment: \{treatment\} \\
Outcome: \{outcome\}
}

\end{tcolorbox}
\caption{Prompt used for ideological expected-sign annotation.}
\label{fig:ideology_prompt}
\end{figure}

\ifPDFTeX
\end{CJK*}
\fi
\end{document}